\documentclass[11pt,a4paper]{article}
\usepackage[hyperref]{emnlp2018}
\usepackage{times}
\usepackage{latexsym}
\usepackage{url}
\usepackage{graphicx}
\usepackage{tabulary}

\aclfinalcopy % Uncomment this line for the final submission

\author{Eben Holderness\textsuperscript{1,2}, Nicholas Miller\textsuperscript{1,2}, Philip Cawkwell\textsuperscript{1}, Kirsten Bolton\textsuperscript{1}, \\ {\bf James Pustejovsky\textsuperscript{2},} {\bf Marie Meteer\textsuperscript{2}} \and {\bf Mei-Hua Hall\textsuperscript{1}} \\
\textsuperscript{1}Psychosis Neurobiology Laboratory, McLean Hospital, Harvard Medical School \\ \textsuperscript{2}Department of Computer Science, Brandeis University \\ {\tt \{eholderness, mhall\}@mclean.harvard.edu} \\ {\tt nicholas.anthony.miller@gmail.com} \\ {\tt \{pcawkwell, kbolton\}@partners.org} \\ {\tt \{jamesp, mmeteer\}@cs.brandeis.edu}\\ }
\title{Analysis of Risk Factor Domains in Psychosis Patient Health Records\newline}

\date{8/31/2018}

\begin{document}
\maketitle
\begin{abstract}
Readmission after discharge from a hospital is disruptive and costly, regardless of the reason. However, it can be particularly problematic for psychiatric patients, so predicting which patients may be readmitted is critically important but also very difficult. Clinical narratives in psychiatric electronic health records (EHRs) span a wide range of topics and vocabulary; therefore, a psychiatric readmission prediction model must begin with a robust and interpretable topic extraction component. We created a data pipeline for using document vector similarity metrics to perform topic extraction on psychiatric EHR data in service of our long-term goal of creating a readmission risk classifier. We show initial results for our topic extraction model and identify additional features we will be incorporating in the future.
\end{abstract}

\section{Introduction}

Psychotic disorders typically emerge in late adolescence or early adulthood \cite{kessler2007age, thomsen1996schizophrenia} and affect approximately 2.5-4\% of the population \cite{perala2007lifetime, bogren2009common}, making them one of the leading causes of disability worldwide \cite{vos2015global}. A substantial proportion of psychiatric inpatients are readmitted after discharge \cite{wiersma1998natural}. Readmissions are disruptive for patients and families, and are a key driver of rising healthcare costs \cite{mangalore2007cost, wu2005economic}. Reducing readmission risk is therefore a major unmet need of psychiatric care. Developing clinically implementable machine learning tools to enable accurate assessment of risk factors associated with readmission offers opportunities to inform the selection of treatment interventions and implement appropriate preventive measures. 

In psychiatry, traditional strategies to study readmission risk factors rely on clinical observation and manual retrospective chart review \cite{olfson1999assessing, lorine2015risk}. This approach, although benefitting from clinical expertise, does not scale well for large data sets, is effort-intensive, and lacks automation. An efficient, more robust, and cheaper NLP-based alternative approach has been developed and met with some success in other medical fields \cite{murff2011automated}. However, this approach has seldom been applied in psychiatry because of the unique characteristics of psychiatric medical record content.

There are several challenges for topic extraction when dealing with clinical narratives in psychiatric EHRs. First, the vocabulary used is highly varied and context-sensitive. A patient may report ``feeling `really great and excited'" -- symptoms of mania -- without any explicit mention of keywords that differ from everyday vocabulary. Also, many technical terms in clinical narratives are multiword expressions (MWEs) such as `obsessive body image', `linear thinking', `short attention span', or `panic attack'. These phrasemes are comprised of words that in isolation do not impart much information in determining relatedness to a given topic but do in the context of the expression. 

Second, the narrative structure in psychiatric clinical narratives varies considerably in how the same phenomenon can be described. Hallucinations, for example, could be described as ``the patient reports auditory hallucinations," or ``the patient has been hearing voices for several months," amongst many other possibilities. 

Third, phenomena can be directly mentioned without necessarily being relevant to the patient specifically. Psychosis patient discharge summaries, for instance, can include future treatment plans (e.g. ``Prevent relapse of a manic or major depressive episode.", ``Prevent recurrence of psychosis.") containing vocabulary that at the word-level seem strongly correlated with readmission risk. Yet at the paragraph-level these do not indicate the presence of a readmission risk factor in the patient and in fact indicate the absence of a risk factor that was formerly present.

Lastly, given the complexity of phenotypic assessment in psychiatric illnesses, patients with psychosis exhibit considerable differences in terms of illness and symptom presentation. The constellation of symptoms leads to various diagnoses and comorbidities that can change over time, including schizophrenia, schizoaffective disorder, bipolar disorder with psychosis, and substance use induced psychosis. Thus, the lexicon of words and phrases used in EHRs differs not only across diagnoses but also across patients and time.

Taken together, these factors make topic extraction a difficult task that cannot be accomplished by keyword search or other simple text-mining techniques.

To identify specific risk factors to focus on, we not only reviewed clinical literature of risk factors associated with readmission \cite{alvarez2012risk, addington2010predictors}, but also considered research related to functional remission \cite{harvey2009toward}, forensic risk factors \cite{singh2010forensic}, and consulted clinicians involved with this project. Seven risk factor domains -- Appearance, Mood, Interpersonal, Occupation, Thought Content, Thought Process, and Substance -- were chosen because they are clinically relevant, consistent with literature, replicable across data sets, explainable, and implementable in NLP algorithms.

In our present study, we evaluate multiple approaches to automatically identify which risk factor domains are associated with which paragraphs in psychotic patient EHRs.\footnote{This study has received IRB approval.} We perform this study in support of our long-term goal of creating a readmission risk classifier that can aid clinicians in targeting individual treatment interventions and assessing patient risk of harm (e.g. suicide risk, homicidal risk). Unlike other contemporary approaches in machine learning, we intend to create a model that is clinically explainable and flexible across training data while maintaining consistent performance. 

To incorporate clinical expertise in the identification of risk factor domains, we undertake an annotation project, detailed in section 3.1. We identify a test set of over 1,600 EHR paragraphs which a team of three domain-expert clinicians annotate paragraph-by-paragraph for relevant risk factor domains. Section 3.2 describes the results of this annotation task. We then use the gold standard from the annotation project to assess the performance of multiple neural classification models trained exclusively on Term Frequency -- Inverse Document Frequency (TF-IDF) vectorized EHR data, described in section 4. To further improve the performance of our model, we incorporate domain-relevant MWEs identified using all in-house data. 

\section{Related Work}
McCoy et al. \shortcite{mccoy2015clinical} constructed a corpus of web data based on the Research Domain Criteria (RDoC)\cite{insel2010research}, and used this corpus to create a vector space document similarity model for topic extraction. They found that the `negative valence' and `social' RDoC domains were associated with readmission. Using web data (in this case data retrieved from the Bing API) to train a similarity model for EHR texts is problematic since it differs from the target data in both structure and content. Based on reconstruction of the procedure, we conclude that many of the informative MWEs critical to understanding the topics of paragraphs in EHRs are not captured in the web data. Additionally, RDoC is by design a generalized research construct to describe the entire spectrum of mental disorders and does not include domains that are based on observation or causes of symptoms. Important indicators within EHRs of patient health, like appearance or occupation, are not included in the RDoC constructs.  

Rumshisky et al. \shortcite{rumshisky2016predicting} used a corpus of EHRs from patients with a primary diagnosis of major depressive disorder to create a 75-topic LDA topic model that they then used in a readmission prediction classifier pipeline. Like with McCoy et al. \shortcite{mccoy2015clinical}, the data used to train the LDA model was not ideal as the generalizability of the data was narrow, focusing on only one disorder. Their model achieved readmission prediction performance with an area under the curve of .784 compared to a baseline of .618. To perform clinical validation of the topics derived from the LDA model, they manually evaluated and annotated the topics, identifying the most informative vocabulary for the top ten topics. With their training data, they found the strongest coherence occurred in topics involving substance use, suicidality, and anxiety disorders. But given the unsupervised nature of the LDA clustering algorithm, the topic coherence they observed is not guaranteed across data sets.

\section{Data}
\footnotetext[2]{The vast majority of patients in our target cohort are \newline dependents on a parental private health insurance plan.}

Our target data set consists of a corpus of discharge summaries, admission notes, individual encounter notes, and other clinical notes from 220 patients in the OnTrack\textsuperscript{TM} program at McLean Hospital. OnTrack\textsuperscript{TM} is an outpatient program, focusing on treating adults ages 18 to 30 who are experiencing their first episodes of psychosis. The length of time in the program varies depending on patient improvement and insurance coverage, with an average of two to three years. The program focuses primarily on early intervention via individual therapy, group therapy, medication evaluation, and medication management. See Table \ref{cohort-data} for a demographic breakdown of the 220 patients, for which we have so far extracted approximately 240,000 total EHR paragraphs spanning from 2011 to 2014 using Meditech, the software employed by McLean for storing and organizing EHR data.

These patients are part of a larger research cohort of approximately 1,800 psychosis patients, which will allow us to connect the results of this EHR study with other ongoing research studies incorporating genetic, cognitive, neurobiological, and functional outcome data from this cohort. 

We also use an additional data set for training our vector space model, comprised of EHR texts queried from the Research Patient Data Registry (RPDR), a centralized regional data repository of clinical data from all institutions in the Partners HealthCare network. These records are highly comparable in style and vocabulary to our target data set. The corpus consists of discharge summaries, encounter notes, and visit notes from approximately 30,000 patients admitted to the system's hospitals with psychiatric diagnoses and symptoms. This breadth of data captures a wide range of clinical narratives, creating a comprehensive foundation for topic extraction. 

After using the RPDR query tool to extract EHR paragraphs from the RPDR database, we created a training corpus by categorizing the extracted paragraphs according to their risk factor domain using a lexicon of 120 keywords that were identified by the clinicians involved in this project. Certain domains -- particularly those involving thoughts and other abstract concepts -- are often identifiable by MWEs rather than single words. The same clinicians who identified the keywords manually examined the bigrams and trigrams with the highest TF-IDF scores for each domain in the categorized paragraphs, identifying those which are conceptually related to the given domain. We then used this lexicon of 775 keyphrases to identify more relevant training paragraphs in RPDR and treat them as (non-stemmed) unigrams when generating the matrix. By converting MWEs such as `shortened attention span', `unusual motor activity', `wide-ranging affect', or `linear thinking' to non-stemmed unigrams, the TF-IDF score (and therefore the predictive value) of these terms is magnified. In total, we constructed a corpus of roughly 100,000 paragraphs consisting of 7,000,000 tokens for training our model.
 
\begin{table}[t!]
\begin{center}
\begin{tabular}{|p{0.7\linewidth} | p{0.2\linewidth}|}
\hline
\bf Mean Age (2014) & 20.7 \\ \hline
\bf Gender (Male) & 79\% \\\hline
\bf Race & \\\hline
\hspace{2.15mm} Asian & 6\% \\\hline
\hspace{2.15mm} Black & 7\% \\\hline
\hspace{2.15mm} Caucasian & 77\% \\\hline
\hspace{2.15mm} Latino & 5\% \\\hline
\hspace{2.15mm} Multiracial & 5\%  \\\hline
\bf Insurance (Public)\footnotemark[2] & 5.5\%\\\hline
\bf 30-day Inpatient Readmission Rate & 14\% \\\hline
\end{tabular}
\end{center}
\caption{Demographic breakdown of the target cohort.}\label{cohort-data}
\end{table}

\begin{table*}
\centering
\small
\begin{tabulary}{0.94\linewidth}{|p{1.5cm}|L|L|L|}
\hline
{\bf Domain} & {\bf  Description} & {\bf Example Paragraph} & {\bf Example Keywords} \\\hline
Appearance & Physical appearance, gestures, and mannerisms & ``A well-appearing, clean young woman appearing her stated age, pleasant and cooperative. Eye contact was good." & disheveled, clothing, groomed, wearing, clean\\\hline  
Thought Content & Suicidal/homicidal ideation, obsessions, phobias, delusions, hallucinations & ``No SI\footnotemark[3], No HI\footnotemark[4], No hallucinations, Ideas of reference, Paranoid delusions" &  obsession, delusion, grandiose, ideation, suicidal, paranoid\\\hline 
Interpersonal & Family situation, friendships, and other social relationships & ``Pt. overall appears to be functioning very well despite this conflict with a romantic interest of hers." & boyfriend, relationship, peers, family, parents, social\\\hline
Mood & Feelings and overall disposition & ``Pt. indicates that his mood is becoming more `depressed.'" & anxious, calm, depressed, labile, confused, cooperative\\\hline
Occupation & School and/or employment & ``Pt. followed through with decision to leave college at this point in time." & boss, employed, job, school, class, homework, work\\\hline
Thought Process & Pace and coherence of thoughts. Includes linear, goal-directed, perseverative, tangential, and flight of ideas & ``Disorganized (Difficult to communicate with patient.), Paucity of thought, Thought-blocking." & linear, tangential, prosody, blocking, goal-directed, perseverant\\\hline
Substance & Drug and/or alcohol use & ``Patient used marijuana once which he believes triggered the current episode." & cocaine, marijuana, ETOH\footnotemark[5], addiction, narcotic\\\hline
Other & Any paragraph that does not fall into any of the other seven domains & ``Maintain mood stabilization, prevent future episodes of mania, improve self-monitoring skills." & --
\\\hline
\end{tabulary}
\caption{Annotation scheme for the domain classification task.}\label{annotation-scheme}
\end{table*}

\subsection{Annotation Task}
In order to evaluate our models, we annotated 1,654 paragraphs selected from the 240,000 paragraphs extracted from Meditech with the clinically relevant domains described in Table \ref{annotation-scheme}. The annotation task was completed by three licensed clinicians. All paragraphs were removed from the surrounding EHR context to ensure annotators were not influenced by the additional contextual information. Our domain classification models consider each paragraph independently and thus we designed the annotation task to mirror the information available to the models.

The annotators were instructed to label each paragraph with one or more of the seven risk factor domains. In instances where more than one domain was applicable, annotators assigned the domains in order of prevalence within the paragraph. An eighth label, `Other', was included if a paragraph was ambiguous, uninterpretable, or about a domain not included in the seven risk factor domains (e.g. non-psychiatric medical concerns and lab results). The annotations were then reviewed by a team of two clinicians who adjudicated collaboratively to create a gold standard. The gold standard and the clinician-identified keywords and MWEs have received IRB approval for release to the community. They are available as supplementary data to this paper.

\subsection{Inter-Annotator Agreement}

\begin{table*}[t!]
\begin{center}
\begin{tabular}{|l|l|l|l|}
\hline \bf Labels & \bf Fleiss's Kappa & \bf Cohen's Multi-Kappa & \bf Mean Accuracy  \\ \hline
Overall & 0.575 & 0.571 & 0.746 \\\hline
First Domain Only & 0.536 & 0.528 & 0.805 \\\hline
\end{tabular}
\end{center}
\caption{Inter-annotator agreement}\label{iaa}
\end{table*}

Inter-annotator agreement (IAA) was assessed using a combination of Fleiss's Kappa (a variant of Scott's Pi that measures pairwise agreement for annotation tasks involving more than two annotators) \cite{fleiss1971measuring} and Cohen's Multi-Kappa as proposed by Davies and Fleiss \shortcite{davies1982measuring}. Table \ref{iaa} shows IAA calculations for both overall agreement and agreement on the first (most important) domain only. Following adjudication, accuracy scores were calculated for each annotator by evaluating their annotations against the gold standard.
 
Overall agreement was generally good and aligned almost exactly with the IAA on the first domain only. Out of the 1,654 annotated paragraphs, 671 (41\%) had total agreement across all three annotators. We defined total agreement for the task as a set-theoretic complete intersection of domains for a paragraph identified by all annotators. 98\% of paragraphs in total agreement involved one domain. Only 35 paragraphs had total disagreement, which we defined as a set-theoretic null intersection between the three annotators. An analysis of the 35 paragraphs with total disagreement showed that nearly 30\% included the term ``blunted/restricted". In clinical terminology, these terms can be used to refer to appearance, affect, mood, or emotion. Because the paragraphs being annotated were extracted from larger clinical narratives and examined independently of any surrounding context, it was difficult for the annotators to determine the most appropriate domain. This lack of contextual information resulted in each annotator using a different `default' label: Appearance, Mood, and Other. During adjudication, Other was decided as the most appropriate label unless the paragraph contained additional content that encompassed other domains, as it avoids making unnecessary assumptions.
\footnotetext[3]{Suicidal ideation}
\footnotetext[4]{Homicidal ideation}
\footnotetext[5]{Ethyl alcohol and ethanol}

\begin{table}[t!]
\begin{center}
\begin{tabular}{|p{0.3\linewidth} | p{0.25\linewidth}| p{0.25\linewidth}|}
\hline
\bf Network & MLP & RBF \\ \hline
 & & \\\hline
\bf Input Layer & & \\\hline
\hspace{2.15mm} Nodes & 100 & 100 \\\hline
\hspace{2.15mm} Dropout & 0.2 & 0.2 \\\hline
\hspace{2.15mm} Activation & ReLU\footnotemark[6] & ReLU \\\hline
\bf Hidden Layer & & \\\hline
\hspace{2.15mm} Nodes & 100 & 350 \\\hline
\hspace{2.15mm} Dropout & 0.5 & 0.0 \\\hline
\hspace{2.15mm} Activation & ReLU & RBF \\\hline
\bf Output Layer & & \\\hline
\hspace{2.15mm} Nodes & 7 & 7 \\\hline
\hspace{2.15mm} Activation & Sigmoid & Linear \\\hline
\bf Optimizer & Adam\footnotemark[7] & Adam \\\hline
\bf Loss Function & Categorical Cross Entropy & Mean Squared Error \\\hline
\bf Training Epochs & 30 & 50 \\\hline
\bf Batch Size & 128 & 128 \\\hline
\end{tabular}
\end{center}
\caption{Architectures of our highest-performing MLP and RBF networks.}\label{architectures}
\end{table}

A Fleiss's Kappa of 0.575 lies on the boundary between `Moderate' and `Substantial' agreement as proposed by Landis and Koch \shortcite{landis1977measurement}. This is a promising indication that our risk factor domains are adequately defined by our present guidelines and can be employed by clinicians involved in similar work at other institutions.

The fourth column in Table \ref{iaa}, Mean Accuracy, was calculated by averaging the three annotator accuracies as evaluated against the gold standard. This provides us with an informative baseline of human parity on the domain classification task.

\footnotetext[6]{Rectified Linear Units, $f(x)=max(0,x)$ \cite{nair2010rectified}}
\footnotetext[7]{Adaptive Moment Estimation \cite{kingma2014adam}}

\section{Topic Extraction}

\begin{figure*}
\begin{center}
\includegraphics[width=\textwidth]{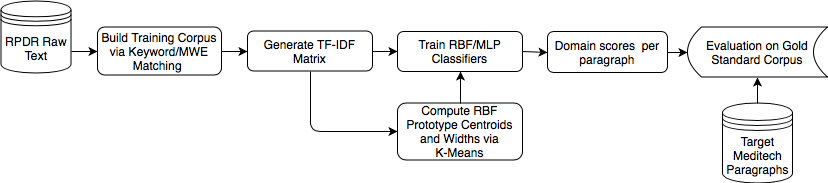}
\caption{Data pipeline for training and evaluating our risk factor domain classifiers.}\label{pipeline}
\end{center}
\end{figure*}

Figure \ref{pipeline} illustrates the data pipeline for generating our training and testing corpora, and applying them to our classification models.

We use the TfidfVectorizer tool included in the scikit-learn machine learning toolkit \cite{pedregosa2011scikit} to generate our TF-IDF vector space models, stemming tokens with the Porter Stemmer tool provided by the NLTK library \cite{bird2004nltk}, and calculating TF-IDF scores for unigrams, bigrams, and trigrams. Applying Singular Value Decomposition (SVD) to the TF-IDF matrix, we reduce the vector space to 100 dimensions, which Zhang et al. \shortcite{zhang2011comparative} found to improve classifier performance.

Starting with the approach taken by McCoy et al. \shortcite{mccoy2015clinical}, who used aggregate cosine similarity scores to compute domain similarity directly from their TF-IDF vector space model, we extend this method by training a suite of three-layer multilayer perceptron (MLP) and radial basis function (RBF) neural networks using a variety of parameters to compare performance. We employ the Keras deep learning library \cite{chollet2015keras} using a TensorFlow backend \cite{abadi2016tensorflow} for this task. The architectures of our highest performing MLP and RBF models are summarized in Table \ref{architectures}. Prototype vectors for the nodes in the hidden layer of our RBF model are selected via k-means clustering \cite{macqueen1967some} on each domain paragraph megadocument individually. The RBF transfer function for each hidden layer node is assigned the same width, which is based off the maximum Euclidean distance between the centroids that were computed using k-means. 

To prevent overfitting to the training data, we utilize a dropout rate \cite{srivastava2014dropout} of 0.2 on the input layer of all models and 0.5 on the MLP hidden layer.

Since our classification problem is multiclass, multilabel, and open-world, we employ seven nodes with sigmoid activations in the output layer, one for each risk factor domain. This allows us to identify paragraphs that fall into more than one of the seven domains, as well as determine paragraphs that should be classified as Other. Unlike the traditionally used softmax activation function, which is ideal for single-label, closed-world classification tasks, sigmoid nodes output class likelihoods for each node independently without the normalization across all classes that occurs in softmax.  

We find that the risk factor domains vary in the degree of homogeneity of language used, and as such certain domains produce higher similarity scores, on average, than others. To account for this, we calculate threshold similarity scores for each domain using the formula min=avg(sim)+$\alpha$*$\sigma$(sim), where $\sigma$ is standard deviation and $\alpha$ is a constant, which we set to 0.78 for our MLP model and 1.2 for our RBF model through trial-and-error. Employing a generalized formula as opposed to manually identifying threshold similarity scores for each domain has the advantage of flexibility in regards to the target data, which may vary in average similarity scores depending on its similarity to the training data. If a paragraph does not meet threshold on any domain, it is classified as Other.  

\section{Results and Discussion}

Table \ref{results} shows the performance of our models on classifying the paragraphs in our gold standard. To assess relative performance of feature representations, we also include performance metrics of our models without MWEs. Because this is a multilabel classification task we use macro-averaging to compute precision, recall, and F1 scores for each paragraph in the testing set. In identifying domains individually, our models achieved the highest per-domain scores on Substance (F1 $\approx$ 0.8) and the lowest scores on Interpersonal and Mood (F1 $\approx$ 0.5). We observe a consistency in per-domain performance rankings between our MLP and RBF models.

\begin{table}[t!]
\begin{tabular}{|p{2.8cm}|p{1.4cm}|p{0.9cm}|p{0.8cm}|}
\hline \bf & \bf Precision & \bf Recall & \bf F1  \\ \hline
Aggregate Cosine Similarity Scores & 0.602 & 0.563 & 0.574 \\\hline
MLP Baseline \newline (No MWEs) & 0.611 & 0.567 & 0.579 \\\hline
RBF Baseline \newline (No MWEs) & 0.603 & 0.618 & 0.606 \\\hline
\bf MLP \newline (w/ MWEs) & 0.717 & 0.666 & 0.681 \\\hline
\hspace{0.3em} Appearance & 0.886 & 0.414 & 0.564 \\\hline
\hspace{0.3em} Interpersonal & 0.548 & 0.453 & 0.496 \\\hline
\hspace{0.3em} Mood & 0.691 & 0.430 & 0.530 \\\hline
\hspace{0.3em} Occupation & 0.826 & 0.461 & 0.592 \\\hline
\hspace{0.3em} Substance & 0.920 & 0.703 & 0.797 \\\hline
\hspace{0.3em} Thought Content & 0.926 & 0.590 & 0.721 \\\hline
\hspace{0.3em} Thought Process & 0.654 & 0.617 & 0.635 \\\hline
\hspace{0.3em} Other & 0.632 & 0.798 & 0.710 \\\hline
\bf RBF \newline (w/ MWEs) & 0.684 & 0.630 & 0.645 \\\hline
\hspace{0.3em} Appearance & 0.670 & 0.490 & 0.566 \\\hline
\hspace{0.3em} Interpersonal & 0.410 & 0.493 & 0.448 \\\hline
\hspace{0.3em} Mood & 0.655 & 0.399 & 0.496 \\\hline
\hspace{0.3em} Occupation & 0.720 & 0.501 & 0.598 \\\hline
\hspace{0.3em} Substance & 0.866 & 0.730 & 0.792 \\\hline
\hspace{0.3em} Thought Content & 0.892 & 0.547 & 0.678 \\\hline
\hspace{0.3em} Thought Process & 0.569 & 0.691 & 0.624 \\\hline
\hspace{0.3em} Other & 0.651 & 0.650 & 0.651 \\\hline

\end{tabular}
\caption{Overall and domain-specific Precision, Recall, and F1 scores for our models. 
The first row computes similarity directly from the TF-IDF matrix, as in \cite{mccoy2015clinical}. All other rows are classifier outputs.}\label{results}
\end{table}

\begin{figure*}
\begin{center}
\includegraphics[width=11.25cm]{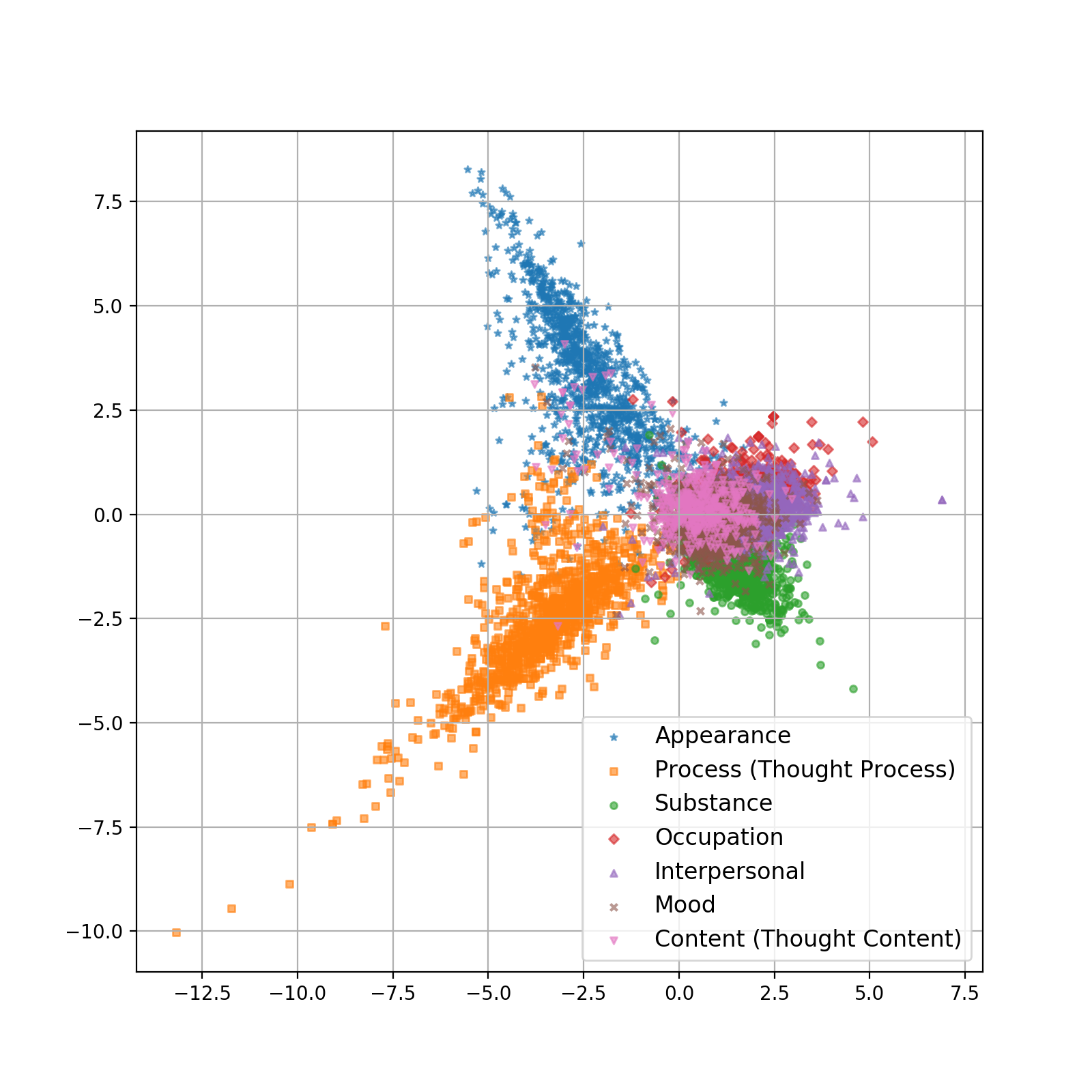}
\caption{2-component linear discriminant analysis of the RPDR training data.}\label{lda}
\end{center}
\end{figure*}
The wide variance in per-domain performance is due to a number of factors. Most notably, the training examples we extracted from RPDR -- while very comparable to our target OnTrack\textsuperscript{TM} data -- may not have an adequate variety of content and range of vocabulary. Although using keyword and MWE matching to create our training corpus has the advantage of being significantly less labor intensive than manually labeling every paragraph in the corpus, it is likely that the homogeneity of language used in the training paragraphs is higher than it would be otherwise. Additionally, all of the paragraphs in the training data are assigned exactly one risk factor domain even if they actually involve multiple risk factor domains, making the clustering behavior of the paragraphs more difficult to define. Figure \ref{lda} illustrates the distribution of paragraphs in vector space using 2-component Linear Discriminant Analysis (LDA) \cite{johnson2004multivariate}.

Despite prior research indicating that similar classification tasks to ours are more effectively performed by RBF networks \cite{scheirer2014probability, jain2014multi, bendale2015towards}, we find that a MLP network performs marginally better with significantly less preprocessing (i.e. k-means and width calculations) involved. We can see in Figure \ref{lda} that Thought Process, Appearance, Substance, and -- to a certain extent -- Occupation clearly occupy specific regions, whereas Interpersonal, Mood, and Thought Content occupy the same noisy region where multiple domains overlap. Given that similarity is computed using Euclidean distance in an RBF network, it is difficult to accurately classify paragraphs that fall in regions occupied by multiple risk factor domain clusters since prototype centroids from the risk factor domains will overlap and be less differentiable. This is confirmed by the results in Table \ref{results}, where the differences in performance between the RBF and MLP models are more pronounced in the three overlapping domains (0.496 vs 0.448 for Interpersonal, 0.530 vs 0.496 for Mood, and 0.721 vs 0.678 for Thought Content) compared to the non-overlapping domains (0.564 vs 0.566 for Appearance, 0.592 vs 0.598 for Occupation, 0.797 vs 0.792 for Substance, and 0.635 vs 0.624 for Thought Process). We also observe a similarity in the words and phrases with the highest TF-IDF scores across the overlapping domains: many of the Thought Content words and phrases with the highest TF-IDF scores involve interpersonal relations (e.g. `fear surrounding daughter', `father', `family history', `familial conflict') and there is a high degree of similarity between high-scoring words for Mood (e.g. `meets anxiety criteria', `cope with mania', `ocd'\footnotemark[8]) and Thought Content (e.g. `mania', `feels anxious', `feels exhausted').

\footnotetext[8]{Obsessive-compulsive disorder}

MWEs play a large role in correctly identifying risk factor domains. Factoring them into our models increased classification performance by ~15\%, a marked improvement over our baseline model. This aligns with our expectations that MWEs comprised of a quotidian vocabulary hold much more clinical significance than when the words in the expressions are treated independently. 

Threshold similarity scores also play a large role in determining the precision and recall of our models: higher thresholds lead to a smaller number of false positives and a greater number of false negatives for each risk factor domain. Conversely, more paragraphs are incorrectly classified as Other when thresholds are set higher. Since our classifier will be used in future work as an early step in a data analysis pipeline for determining readmission risk, misclassifying a paragraph with an incorrect risk factor domain at this stage can lead to greater inaccuracies at later stages. Paragraphs misclassified as Other, however, will be discarded from the data pipeline. Therefore, we intentionally set a conservative threshold where only the most confidently labeled paragraphs are assigned membership in a particular domain.  

\section{Future Work and Conclusion}
To achieve our goal of creating a framework for a readmission risk classifier, the present study performed necessary evaluation steps by updating and adding to our model iteratively. In the first stage of the project, we focused on collecting the data necessary for training and testing, and on the domain classification annotation task. At the same time, we began creating the tools necessary for automatically extracting domain relevance scores at the paragraph and document level from patient EHRs using several forms of vectorization and topic modeling. In future versions of our risk factor domain classification model we will explore increasing robustness through sequence modeling that considers more contextual information.

Our current feature set for training a machine learning classifier is relatively small, consisting of paragraph domain scores, bag-of-words, length of stay, and number of previous admissions, but we intend to factor in many additional features that extend beyond the scope of the present study. These include a deeper analysis of clinical narratives in EHRs: our next task will be to extend our EHR data pipeline by distinguishing between clinically positive and negative phenomena within each risk factor domain. This will involve a series of annotation tasks that will allow us to generate lexicon-based and corpus-based sentiment analysis tools. We can then use these clinical sentiment scores to generate a gradient of patient improvement or deterioration over time. 

We will also take into account structured data that have been collected on the target cohort throughout the course of this study such as brain based electrophysiological (EEG) biomarkers, structural brain anatomy from MRI scans (gray matter volume, cortical thickness, cortical surface-area), social and role functioning assessments, personality assessment (NEO-FFI\footnotemark[9]), and various symptom scales (PANSS\footnotemark[10], MADRS\footnotemark[11], YMRS\footnotemark[12]). For each feature we consider adding, we will evaluate the performance of the classifier with and without the feature to determine its contribution as a predictor of readmission.

\section{Acknowledgments}
This work was supported by a grant from the National Institute of Mental Health (grant no. 5R01MH109687 to Mei-Hua Hall). We would also like to thank the LOUHI 2018 Workshop reviewers for their constructive and helpful comments. 

\footnotetext[9]{NEO Five-Factor Inventory \cite{costa2010neo}}
\footnotetext[10]{Positive and Negative Syndrome Scale \cite{kay1987positive}}
\footnotetext[11]{Montgomery-Asperg Depression Rating Scale \cite{montgomery1979new}}
\footnotetext[12]{Young Mania Rating Scale \cite{young1978rating}}
% \bibliography{emnlp2018}

\begin{thebibliography}{38}
\expandafter\ifx\csname natexlab\endcsname\relax\def\natexlab#1{#1}\fi

\bibitem[{Abadi et~al.()Abadi, Barham, Chen, Chen, Davis, Dean, Devin,
  Ghemawat, Irving, Isard et~al.}]{abadi2016tensorflow}
Mart{\'\i}n Abadi, Paul Barham, Jianmin Chen, Zhifeng Chen, Andy Davis, Jeffrey
  Dean, Matthieu Devin, Sanjay Ghemawat, Geoffrey Irving, Michael Isard, et~al.
\newblock Tensorflow: a system for large-scale machine learning.

\bibitem[{Addington et~al.(2010)Addington, Beck, Wang, Adams, Pryce, Zhu, Kang,
  and McKenzie}]{addington2010predictors}
Donald~Emile Addington, Cindy Beck, JianLi Wang, Beverly Adams, Cathy Pryce,
  Haifeng Zhu, Jian Kang, and Emily McKenzie. 2010.
\newblock Predictors of admission in first-episode psychosis: developing a risk
  adjustment model for service comparisons.
\newblock \emph{Psychiatric Services}, 61(5):483--488.

\bibitem[{Alvarez-Jimenez et~al.(2012)Alvarez-Jimenez, Priede, Hetrick,
  Bendall, Killackey, Parker, McGorry, and Gleeson}]{alvarez2012risk}
Mario Alvarez-Jimenez, A~Priede, SE~Hetrick, Sarah Bendall, Eoin Killackey,
  AG~Parker, PD~McGorry, and JF~Gleeson. 2012.
\newblock Risk factors for relapse following treatment for first episode
  psychosis: a systematic review and meta-analysis of longitudinal studies.
\newblock \emph{Schizophrenia Research}, 139(1-3):116--128.

\bibitem[{Bendale and Boult(2015)}]{bendale2015towards}
Abhijit Bendale and Terrance Boult. 2015.
\newblock Towards open world recognition.
\newblock In \emph{Proceedings of the IEEE Conference on Computer Vision and
  Pattern Recognition}, pages 1893--1902.

\bibitem[{Bird et~al.(2009)Bird, Klein, and Loper}]{bird2004nltk}
Steven Bird, Ewan Klein, and Edward Loper. 2009.
\newblock \emph{Natural language processing with Python: analyzing text with
  the natural language toolkit}.
\newblock " O'Reilly Media, Inc.".

\bibitem[{Bogren et~al.(2009)Bogren, Mattisson, Isberg, and
  Nettelbladt}]{bogren2009common}
Mats Bogren, Cecilia Mattisson, Per-Erik Isberg, and Per Nettelbladt. 2009.
\newblock How common are psychotic and bipolar disorders? a 50-year follow-up
  of the lundby population.
\newblock \emph{Nordic journal of psychiatry}, 63(4):336--346.

\bibitem[{Chollet et~al.(2015)}]{chollet2015keras}
Fran\c{c}ois Chollet et~al. 2015.
\newblock Keras.
\newblock \url{https://keras.io}.

\bibitem[{Costa and McCrae(2010)}]{costa2010neo}
PT~Costa and Robert~R McCrae. 2010.
\newblock The neo personality inventory: 3.
\newblock \emph{Odessa, FL: Psychological assessment resources}.

\bibitem[{Davies and Fleiss(1982)}]{davies1982measuring}
Mark Davies and Joseph~L Fleiss. 1982.
\newblock Measuring agreement for multinomial data.
\newblock \emph{Biometrics}, pages 1047--1051.

\bibitem[{Fleiss(1971)}]{fleiss1971measuring}
Joseph~L Fleiss. 1971.
\newblock Measuring nominal scale agreement among many raters.
\newblock \emph{Psychological bulletin}, 76(5):378.

\bibitem[{Harvey and Bellack(2009)}]{harvey2009toward}
Philip~D Harvey and Alan~S Bellack. 2009.
\newblock Toward a terminology for functional recovery in schizophrenia: is
  functional remission a viable concept?
\newblock \emph{Schizophrenia Bulletin}, 35(2):300--306.

\bibitem[{Insel et~al.(2010)Insel, Cuthbert, Garvey, Heinssen, Pine, Quinn,
  Sanislow, and Wang}]{insel2010research}
Thomas Insel, Bruce Cuthbert, Marjorie Garvey, Robert Heinssen, Daniel~S Pine,
  Kevin Quinn, Charles Sanislow, and Philip Wang. 2010.
\newblock Research domain criteria (rdoc): toward a new classification
  framework for research on mental disorders.

\bibitem[{Jain et~al.(2014)Jain, Scheirer, and Boult}]{jain2014multi}
Lalit~P Jain, Walter~J Scheirer, and Terrance~E Boult. 2014.
\newblock Multi-class open set recognition using probability of inclusion.
\newblock In \emph{European Conference on Computer Vision}, pages 393--409.
  Springer.

\bibitem[{Johnson and Wichern(2004)}]{johnson2004multivariate}
Richard~A Johnson and Dean~W Wichern. 2004.
\newblock Multivariate analysis.
\newblock \emph{Encyclopedia of Statistical Sciences}, 8.

\bibitem[{Kay et~al.(1987)Kay, Fiszbein, and Opler}]{kay1987positive}
Stanley~R Kay, Abraham Fiszbein, and Lewis~A Opler. 1987.
\newblock The positive and negative syndrome scale (panss) for schizophrenia.
\newblock \emph{Schizophrenia bulletin}, 13(2):261--276.

\bibitem[{Kessler et~al.(2007)Kessler, Amminger, Aguilar-Gaxiola, Alonso, Lee,
  and Ustun}]{kessler2007age}
Ronald~C Kessler, G~Paul Amminger, Sergio Aguilar-Gaxiola, Jordi Alonso, Sing
  Lee, and T~Bedirhan Ustun. 2007.
\newblock Age of onset of mental disorders: a review of recent literature.
\newblock \emph{Current opinion in psychiatry}, 20(4):359.

\bibitem[{Kingma and Ba(2014)}]{kingma2014adam}
Diederik~P Kingma and Jimmy Ba. 2014.
\newblock Adam: A method for stochastic optimization.
\newblock \emph{arXiv preprint arXiv:1412.6980}.

\bibitem[{Landis and Koch(1977)}]{landis1977measurement}
J~Richard Landis and Gary~G Koch. 1977.
\newblock The measurement of observer agreement for categorical data.
\newblock \emph{biometrics}, pages 159--174.

\bibitem[{Lorine et~al.(2015)Lorine, Goenjian, Kim, Steinberg, Schmidt, and
  Goenjian}]{lorine2015risk}
Kim Lorine, Haig Goenjian, Soeun Kim, Alan~M Steinberg, Kendall Schmidt, and
  Armen~K Goenjian. 2015.
\newblock Risk factors associated with psychiatric readmission.
\newblock \emph{The Journal of nervous and mental disease}, 203(6):425--430.

\bibitem[{MacQueen et~al.(1967)}]{macqueen1967some}
James MacQueen et~al. 1967.
\newblock Some methods for classification and analysis of multivariate
  observations.

\bibitem[{Mangalore and Knapp(2007)}]{mangalore2007cost}
Roshni Mangalore and Martin Knapp. 2007.
\newblock Cost of schizophrenia in england.
\newblock \emph{The journal of mental health policy and economics},
  10(1):23--41.

\bibitem[{McCoy et~al.(2015)McCoy, Castro, Rosenfield, Cagan, Kohane, and
  Perlis}]{mccoy2015clinical}
Thomas~H McCoy, Victor~M Castro, Hannah~R Rosenfield, Andrew Cagan, Isaac~S
  Kohane, and Roy~H Perlis. 2015.
\newblock A clinical perspective on the relevance of research domain criteria
  in electronic health records.
\newblock \emph{American Journal of Psychiatry}, 172(4):316--320.

\bibitem[{Montgomery and {\AA}sberg(1979)}]{montgomery1979new}
Stuart~A Montgomery and MARIE {\AA}sberg. 1979.
\newblock A new depression scale designed to be sensitive to change.
\newblock \emph{The British journal of psychiatry}, 134(4):382--389.

\bibitem[{Murff et~al.(2011)Murff, FitzHenry, Matheny, Gentry, Kotter, Crimin,
  Dittus, Rosen, Elkin, Brown et~al.}]{murff2011automated}
Harvey~J Murff, Fern FitzHenry, Michael~E Matheny, Nancy Gentry, Kristen~L
  Kotter, Kimberly Crimin, Robert~S Dittus, Amy~K Rosen, Peter~L Elkin,
  Steven~H Brown, et~al. 2011.
\newblock Automated identification of postoperative complications within an
  electronic medical record using natural language processing.
\newblock \emph{Jama}, 306(8):848--855.

\bibitem[{Nair and Hinton(2010)}]{nair2010rectified}
Vinod Nair and Geoffrey~E Hinton. 2010.
\newblock Rectified linear units improve restricted boltzmann machines.
\newblock In \emph{Proceedings of the 27th international conference on machine
  learning (ICML-10)}, pages 807--814.

\bibitem[{Olfson et~al.(1999)Olfson, Mechanic, Boyer, Hansell, Walkup, and
  Weiden}]{olfson1999assessing}
Mark Olfson, David Mechanic, Carol~A Boyer, Stephen Hansell, James Walkup, and
  Peter~J Weiden. 1999.
\newblock Assessing clinical predictions of early rehospitalization in
  schizophrenia.
\newblock \emph{The Journal of nervous and mental disease}, 187(12):721--729.

\bibitem[{Pedregosa et~al.(2011)Pedregosa, Varoquaux, Gramfort, Michel,
  Thirion, Grisel, Blondel, Prettenhofer, Weiss, Dubourg
  et~al.}]{pedregosa2011scikit}
Fabian Pedregosa, Ga{\"e}l Varoquaux, Alexandre Gramfort, Vincent Michel,
  Bertrand Thirion, Olivier Grisel, Mathieu Blondel, Peter Prettenhofer, Ron
  Weiss, Vincent Dubourg, et~al. 2011.
\newblock Scikit-learn: Machine learning in python.
\newblock \emph{Journal of machine learning research}, 12(Oct):2825--2830.

\bibitem[{Per{\"a}l{\"a} et~al.(2007)Per{\"a}l{\"a}, Suvisaari, Saarni,
  Kuoppasalmi, Isomets{\"a}, Pirkola, Partonen, Tuulio-Henriksson, Hintikka,
  Kiesepp{\"a} et~al.}]{perala2007lifetime}
Jonna Per{\"a}l{\"a}, Jaana Suvisaari, Samuli~I Saarni, Kimmo Kuoppasalmi,
  Erkki Isomets{\"a}, Sami Pirkola, Timo Partonen, Annamari Tuulio-Henriksson,
  Jukka Hintikka, Tuula Kiesepp{\"a}, et~al. 2007.
\newblock Lifetime prevalence of psychotic and bipolar i disorders in a general
  population.
\newblock \emph{Archives of general psychiatry}, 64(1):19--28.

\bibitem[{Rumshisky et~al.(2016)Rumshisky, Ghassemi, Naumann, Szolovits,
  Castro, McCoy, and Perlis}]{rumshisky2016predicting}
A~Rumshisky, M~Ghassemi, T~Naumann, P~Szolovits, VM~Castro, TH~McCoy, and
  RH~Perlis. 2016.
\newblock Predicting early psychiatric readmission with natural language
  processing of narrative discharge summaries.
\newblock \emph{Translational psychiatry}, 6(10):e921.

\bibitem[{Scheirer et~al.(2014)Scheirer, Jain, and
  Boult}]{scheirer2014probability}
Walter~J Scheirer, Lalit~P Jain, and Terrance~E Boult. 2014.
\newblock Probability models for open set recognition.
\newblock \emph{IEEE transactions on pattern analysis and machine
  intelligence}, 36(11):2317--2324.

\bibitem[{Singh and Fazel(2010)}]{singh2010forensic}
Jay~P Singh and Seena Fazel. 2010.
\newblock Forensic risk assessment: A metareview.
\newblock \emph{Criminal Justice and Behavior}, 37(9):965--988.

\bibitem[{Srivastava et~al.(2014)Srivastava, Hinton, Krizhevsky, Sutskever, and
  Salakhutdinov}]{srivastava2014dropout}
Nitish Srivastava, Geoffrey Hinton, Alex Krizhevsky, Ilya Sutskever, and Ruslan
  Salakhutdinov. 2014.
\newblock Dropout: a simple way to prevent neural networks from overfitting.
\newblock \emph{The Journal of Machine Learning Research}, 15(1):1929--1958.

\bibitem[{Thomsen(1996)}]{thomsen1996schizophrenia}
PH~Thomsen. 1996.
\newblock Schizophrenia with childhood and adolescent onset—a nationwide
  register-based study.
\newblock \emph{Acta Psychiatrica Scandinavica}, 94(3):187--193.

\bibitem[{Vos et~al.(2015)Vos, Barber, Bell, Bertozzi-Villa, Biryukov,
  Bolliger, Charlson, Davis, Degenhardt, Dicker et~al.}]{vos2015global}
Theo Vos, Ryan~M Barber, Brad Bell, Amelia Bertozzi-Villa, Stan Biryukov, Ian
  Bolliger, Fiona Charlson, Adrian Davis, Louisa Degenhardt, Daniel Dicker,
  et~al. 2015.
\newblock Global, regional, and national incidence, prevalence, and years lived
  with disability for 301 acute and chronic diseases and injuries in 188
  countries, 1990--2013: a systematic analysis for the global burden of disease
  study 2013.
\newblock \emph{The Lancet}, 386(9995):743--800.

\bibitem[{Wiersma et~al.(1998)Wiersma, Nienhuis, Slooff, and
  Giel}]{wiersma1998natural}
Durk Wiersma, Fokko~J Nienhuis, Cees~J Slooff, and Robert Giel. 1998.
\newblock Natural course of schizophrenic disorders: a 15-year followup of a
  dutch incidence cohort.
\newblock \emph{Schizophrenia bulletin}, 24(1):75--85.

\bibitem[{Wu et~al.(2005)Wu, Birnbaum, Shi, Ball, Kessler, Moulis, and
  Aggarwal}]{wu2005economic}
Eric~Q Wu, Howard~G Birnbaum, Lizheng Shi, Daniel~E Ball, Ronald~C Kessler,
  Matthew Moulis, and Jyoti Aggarwal. 2005.
\newblock The economic burden of schizophrenia in the united states in 2002.
\newblock \emph{Journal of Clinical Psychiatry}, 66(9):1122--1129.

\bibitem[{Young et~al.(1978)Young, Biggs, Ziegler, and Meyer}]{young1978rating}
RC~Young, JT~Biggs, VE~Ziegler, and DA~Meyer. 1978.
\newblock A rating scale for mania: reliability, validity and sensitivity.
\newblock \emph{The British Journal of Psychiatry}, 133(5):429--435.

\bibitem[{Zhang et~al.(2011)Zhang, Yoshida, and Tang}]{zhang2011comparative}
Wen Zhang, Taketoshi Yoshida, and Xijin Tang. 2011.
\newblock A comparative study of tf* idf, lsi and multi-words for text
  classification.
\newblock \emph{Expert Systems with Applications}, 38(3):2758--2765.

\end{thebibliography}

\bibliographystyle{acl_natbib_nourl}

\end{document}